\documentclass{article}

\usepackage{microtype}
\usepackage{graphicx}
\usepackage{subcaption}
\usepackage{booktabs} %

\usepackage{hyperref}

\usepackage[preprint]{icml2026}

\usepackage{amsmath}
\usepackage{amssymb}
\usepackage{mathtools}
\usepackage{amsthm}

\usepackage[capitalize,noabbrev]{cleveref}

\theoremstyle{plain}

\theoremstyle{definition}

\theoremstyle{remark}

\usepackage[textsize=tiny]{todonotes}

\newcommand{\rparagraph}[1]{\vspace{1.2mm}\noindent\textbf{#1.}}

\newcommand{\sparagraph}[1]{\vspace{0.0mm}\noindent\textbf{#1.}}

\newcommand{\ours}{{\textsc{Inspo}}}

\definecolor{Gray}{gray}{0.92}

\definecolor{ourred}{HTML}{F19C99}
\definecolor{ourblue}{HTML}{7EA6E0}
\definecolor{plotred}{HTML}{f77189}
\definecolor{plotgreen}{HTML}{33b07a}
\definecolor{plotpurple}{HTML}{cc7af4}
\definecolor{plotmagenta}{HTML}{f565cc}
\definecolor{plotazure}{HTML}{38a9c5}

\definecolor{codegreen}{rgb}{0,0.6,0}
\definecolor{codegray}{rgb}{0.5,0.5,0.5}
\definecolor{codepurple}{rgb}{0.58,0,0.82}
\definecolor{backcolour}{rgb}{0.95,0.95,0.92}

\definecolor{stage1}{HTML}{34A853}
\definecolor{stage2}{HTML}{A680B8}
\definecolor{stage3}{HTML}{009999}

\definecolor{paleblue}{HTML}{7EA6E0}
\definecolor{palered}{HTML}{F19C99}

\definecolor{mainboxbg}{HTML}{F7F9FC}
\definecolor{mainboxborder}{HTML}{A1C6EA}
\definecolor{mainboxbg2}{HTML}{FFDEDE}
\definecolor{mainboxborder2}{HTML}{FF9393}

\definecolor{racing-green}{RGB}{0, 185, 0}
\definecolor{awesome-red}{RGB}{255, 94, 94}
\definecolor{g-red}{RGB}{213, 66, 56}
\definecolor{g-green}{RGB}{49, 149, 79}
\definecolor{g-purple}{RGB}{181, 116, 157}
\definecolor{g-brown}{RGB}{232, 140, 79}

\definecolor{wan-brown}{RGB}{255, 245, 235}

\definecolor{wan-purple}{RGB}{229, 229, 255}

\definecolor{g-grey}{RGB}{153, 153, 153}

\usepackage{listings}

\lstdefinestyle{mystyle}{
    backgroundcolor=\color{backcolour},   
    commentstyle=\color{codegreen},
    keywordstyle=\color{magenta},
    numberstyle=\tiny\color{codegray},
    stringstyle=\color{codepurple},
    basicstyle=\ttfamily\scriptsize,
    breakatwhitespace=false,         
    breaklines=true,                 
    captionpos=b,                    
    keepspaces=true,                 
    numbersep=5pt,                  
    showspaces=false,                
    showstringspaces=false,
    showtabs=false,                  
    tabsize=2,
    frame=none,
    aboveskip=1pt,
    belowskip=1pt,
}
\usepackage{tcolorbox}

\usepackage{colortbl}
\usepackage{float}
\usepackage{multirow}
\usepackage{wrapfig}
\usepackage{booktabs}
\usepackage{algorithm}
\usepackage{algorithmic}
\usepackage{enumitem}

\icmltitlerunning{Agentic Policy Optimization via Instruction-Policy Co-Evolution}

\begin{document}

\twocolumn[
  \icmltitle{Agentic Policy Optimization via Instruction-Policy Co-Evolution}

  \icmlsetsymbol{equal}{*}
  \icmlsetsymbol{equal}{*}

  \begin{icmlauthorlist}
    \icmlauthor{Han Zhou}{1}
    \icmlauthor{Xingchen Wan}{2,equal}
    \icmlauthor{Ivan Vuli\'c}{1}
    \icmlauthor{Anna Korhonen}{1}
  \end{icmlauthorlist}

  \icmlaffiliation{1}{Language Technology Lab, University of Cambridge.}
  \icmlaffiliation{2}{Machine Learning Research Group, University of Oxford. $^*$Now at Google}

  \icmlcorrespondingauthor{Han Zhou}{hz416@cam.ac.uk}

  \vskip 0.3in
]

\printAffiliationsAndNotice{}  %

\begin{abstract}
Reinforcement Learning with Verifiable Rewards (RLVR) has advanced the reasoning capability of large language models (LLMs), enabling autonomous agents that can conduct effective multi-turn and tool-integrated reasoning. While instructions serve as the primary protocol for defining agents, RLVR typically relies on static and manually designed instructions. However, those instructions may be suboptimal for the base model, and the optimal instruction may change as the agent's policy improves and explores the interaction with the environment. To bridge the gap, we introduce \ours, a novel \textbf{Ins}truction-\textbf{po}licy Co-Evolution framework that integrates instruction optimization as a dynamic component of the reinforcement learning (RL) loop. \ours~maintains a dynamic population of instruction candidates that are sampled with questions, where reward signals in RL loops are automatically attributed to each instruction, and low performers are periodically pruned. New instructions are generated and verified through an on-policy reflection mechanism, where an LLM-based optimizer analyzes past experience from a replay buffer and evolves more effective strategies given the current policy. We conduct extensive experiments on multi-turn retrieval and reasoning tasks, demonstrating that \ours~substantially outperforms strong baselines relying on static instructions. \ours~discovers innovative instructions that guide the agent toward more strategic reasoning paths, achieving substantial performance gains with only a marginal increase in computational overhead.
\end{abstract}

\section{Introduction}
The advent of large language models (LLMs) \citep{brown2020language, chung2024scaling}
has given rise to autonomous agents that are capable of reasoning, interpreting user intents, and tackling complex tasks via interacting with the environment \citep{yao2023react}. When paired with carefully engineered instructions, LLM-based agents have excelled in a wide range of applications, such as code generation \citep{jimenez2023swe}, retrieval-augmented generation \citep{trivedi-etal-2023-interleaving}, and interactive decision-making \citep{su2025learnbyinteract}. Recently, the reinforcement learning (RL) \citep{sutton1999reinforcement} paradigm has further advanced the reasoning capabilities of LLM agents, enabling them to learn policies from verifiable rewards \citep{shao2024deepseekmath} (RLVR) and achieve multi-turn and tool-integrated reasoning \citep{jin2025search, xue2025simpletir}.

In the core of these agentic capabilities, \textit{instructions} serve as the protocol for programming these agents, characterizing their roles, and defining any available tools/interfaces for interaction. The performance of LLM-based agents has been shown to be highly dependent on the instruction \citep{zhou2025multi}, and subtle changes can exert substantial differences in generated trajectories, preventing robust and generalizable agent applications. The compounding effect of instructions is further amplified when LLMs are post-trained via RL, where changes in instructions can result in different initial spaces for policy learning, thereby largely affecting the converged performance after training \citep{liu2025understanding}. Consequently, instruction design becomes crucial for agent training and typically requires costly human efforts for iterative refinements via trial-and-error. 

The traditional paradigm of RLVR treats instruction as a \textit{static} and \textit{pre-defined} input. However, the optimal instruction for the base model is not always known \textit{a priori} and may even change as the model's policy improves and explores the interaction with the environment \citep{soylu-etal-2024-fine}. Recent findings also underscore the importance of instruction for RL, where injecting reward specification \citep{zhang2025merf} or in-context hints \citep{liu2025ghpo} into the instruction better aligns the model with the learning objective and generates richer reward signals. While automated prompt optimization (APO) \citep{zhou2023large, yang2024large} approaches exist for obtaining a better instruction before commencing the RL phase, generalizing them to the online setting of RL and incorporating adaptive knowledge during policy updates is rather non-trivial. 

To bridge this gap, we propose to automate instruction learning not as a static term, but as an integral and dynamic component of the RL learning loop, allowing the instruction and policy to co-evolve in an online setup. We introduce \ours, \textsc{Ins}truction-\textsc{po}licy co-evolution, for agentic policy optimization, a novel framework that delivers two major innovations: 

\begin{itemize}[leftmargin=5mm]
    \item \textit{Policy-Optimization with Dynamic Instructions}: In contrast to existing RLVR approaches that rely on a static instruction, \ours~maintains a dynamic population of instruction candidates. During policy optimization, instructions are sampled from this population and weighted by their importance. The reward signals not only provide the policy gradients, but also attribute reward signals to each instruction candidate based on the assignment. We periodically prune low-performing candidates via successive halving and evolve new instructions from high-performing ones, ensuring a self-improving set of instructions that learns to incorporate new knowledge for policy learning.
    \item \textit{Experience-Driven Instruction Generation}: To generate innovative and effective instruction candidates, we introduce an on-policy reflection mechanism that mimics the idea of experience replay. \ours~utilizes a replay buffer that prioritizes past trajectories where the agent made incorrect or low-reward predictions. The history of experience then provides a reflection signal to an LLM-based instruction-proposer module, which analyzes the failure cases and evolves new instructions via self-reflection. These evolved instructions explicitly guide the agent to avoid previous failure modes and discover innovative new strategies. Lastly, the newly proposed instructions are verified via a low-cost proxy before merging into the active population, where low-performing candidates are rejected. This step ensures a self-improving population of instructions in evolution.
\end{itemize}

We validate our approach through extensive experiments on multi-turn retrieval and reasoning tasks. The experimental results show that \ours~substantially outperforms strong baselines that use static instructions. We demonstrate that \ours~evolves precise and innovative instructions that guide the LLM to discover more strategic reasoning paths and avoid erroneous patterns. Our ablation study further confirms that the online co-evolution of instruction and policy outperforms optimizing the instruction either before or after the RL phase. By enabling the online automation of instruction optimization alongside the policy optimization, \ours~incurs only a marginal computation overhead to the training while bringing substantial performance gains. 

\section{Preliminaries}

\subsection{Reinforcement Learning with Tool-Using}
We formalize multi-turn tool-using as a Markov Decision Process (MDP) \citep{hauskrecht2013hierarchical}, where the LLM-based agent, governed by its policy $\pi$ and parameterized by the weights $\theta$, interacts with an external environment by calling a tool engine $\mathcal{T}$. 
 It generates a trajectory $\tau$ as its state that comprises multiple turns of interactions:  $\tau=(I,q, a_{0},o_{0},\dots,a_{k-1},o_{k-1},a_{k})$, where $I$ is the initial instruction and $q$ is the question input. $a_k$ is the response from the LLM given the history trajectory up to turn $k$, expressing the action of the agent for each turn, such that $a \sim \pi_{\theta}(\cdot | I,q;T)$. The observation $o$ is the feedback from the environment by calling the tool engine, $o_k=\mathcal{T}(a_k)$, which is concatenated in an interleaved manner with the action that forms the final trajectory. 

 The learning objective of RL is to optimize the policy parameters $\theta$ by maximizing the expected returns over a distribution of questions $\mathcal{D}$, where $r(\tau)$ defines the reward function. In RLVR, the reward is typically defined via a rule-based verifier, such as an exact match reward function, $r_{\text{EM}}(\tau)\in\{1, 0\}$, which returns $1$ only if the final action matches the ground truth, and $0$ otherwise. Therefore, the learning objective of RL is expressed as:
\begin{equation}
    \max_{\pi_\theta} \mathbb{E}_{q \sim \mathcal{D}, \tau \sim \pi_\theta(\cdot|I, q;\mathcal{T})} \left[ r(\tau) 
    \right].
\end{equation}

\begin{figure*}[!t]
    \centering
    \includegraphics[width=1\linewidth]{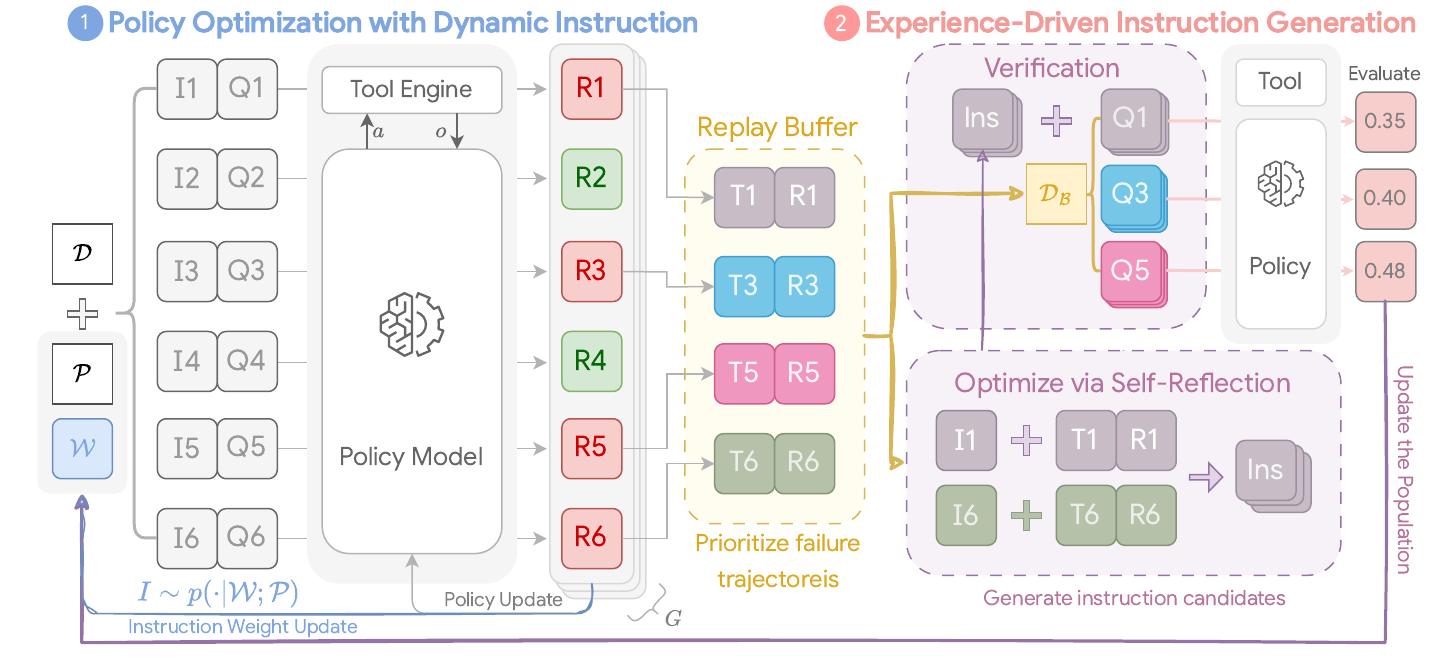}
    \vspace{-4mm}
    \caption{Illustration of \ours: In \textcolor{paleblue}{phase 1}, \ours~maintains a dynamic population of instruction candidates. For each sampled question, the instruction is sampled based on a selection probability weighted by the importance of each instruction. The reward signals derived from the responses here not only update the policy model but also update the importance of instructions. In addition, \ours~involves a replay buffer that prioritizes failure trajectories or trajectories that come with low rewards (marked in \textcolor{g-red}{red}) for experience-driven self-reflection. In \textcolor{palered}{phase 2}, the history of experience then provides rich correction signals to an LLM-based instruction-proposer module, which analyzes the failure cases and evolves new instructions via self-reflection. New instructions are then passed for verification, where the top-performing candidates are merged into the active population of instructions, maintaining a self-improving population of instructions.}
    \label{fig:main}
\end{figure*}
\subsection{Group Relative Policy Optimization (GRPO)}
Group Relative Policy Optimization (GRPO) \citep{shao2024deepseekmath} is a state-of-the-art policy optimization method for language model that derives from Proximal Policy Optimization (PPO) \citep{schulman2017proximal} and obviates the need for a separate model for value function approximation, but instead uses the group-wise sampling to estimate an advantage function $A$ based on relative performance within a group $G$. For each pair of instruction $I$ and question $q\sim\mathcal{D}$, GRPO samples a group of trajectories $\{\tau_i\}_{i=1}^G$ with tool interaction from the old policy $\pi_{\theta_{\text{old}}}(\cdot \mid I, q; \mathcal{T})$ and optimizes the following objective with a KL divergence $\mathbb{D}_\text{KL}\left( \pi_\theta \,||\, \pi_{\text{ref}} \right)$ penalty with the reference policy $\pi_{\text{ref}}$ for regularization:
\begin{equation}
\begin{aligned}
&\mathcal{J}_{\text{GRPO}}(\theta) = \mathbb{E}_{\tau \sim \pi_{\text{old}}} \Bigg[ \frac{1}{G} \sum_{i=1}^G \frac{1}{\sum_t \mathcal{I}_{i,t}} \sum_{t=1}^{|\tau_i|}\mathcal{I}_{i,t} \\
&\quad \cdot \min \left( \rho_{i,t}\hat{A}_{i,t}, \text{clip}(\rho_{i,t}, 1-\epsilon, 1+\epsilon) \hat{A}_{i,t} \right)  - \beta \mathbb{D}_{\text{KL}} \Bigg],
\end{aligned}
\label{eq1}
\end{equation}
where $\rho_{i,t}(\theta)$ is the token-level importance ratio, and $\hat{A}_i$ is the advantage that computes the relative rewards normalized in a group of outputs $\{r_j\}_{j=1}^{G}$:
\begin{equation}
    \rho_{i,t}(\theta) = \frac{\pi_\theta(\tau_{i,t}|\tau_{i,<t})}{\pi_{\theta_{\text{old}}}(\tau_{i,t}|\tau_{i,<t})}, 
    \quad 
    \hat{A}_i = \frac{r_i - \text{mean}(\{r_j\}_{j=1}^{G})}{\text{std}(\{r_j\}_{j=1}^{G})}.
\end{equation}
Different to training a single-turn agent, applying policy gradients to the complete trajectory with environmental observations $o$ that are off-policy may make the training process unstable and result in unintended reward collapse \citep{jin2025search, xue2025simpletir}. Hence, as shown in Eq.~(\ref{eq1}), GRPO for multi-turn RL involves an indicator function, $\mathcal{I}_{i,t}$, that masks out any token-level responses that are from the observation $o$, and the loss is computed only over tokens in $a$ that are from the agent's own actions. 

\section{Instruction-Policy Co-Evolution (\ours)}

The conventional RLVR approach relies on a static and pre-defined instruction, which may be suboptimal and fails to adapt to observations from the environment as the agent's policy updates, potentially constraining the exploration of the agent and resulting in a local optimum for policy learning. Instead, \ours~re-frames the instruction design in RLVR not as a one-off human engineering, but as a dynamic and integral component of the RL loop that can be automatically optimized. We introduce \ours, \textbf{Ins}truction-\textbf{po}licy co-evolution, a novel co-evolving framework where the instruction is optimized alongside the agent's policy learning, enabling the discovery of more effective strategies and reasoning pathways in an online setting. The core of \ours~is composed of two synergistic components: 1) policy optimization with a dynamic set of instructions; 2) an experience-driven self-reflection mechanism for generating new instructions based on the current policy.

\subsection{Policy Optimization with Dynamic Instructions}
To encourage the exploration of agents with different reasoning strategies, \ours~first maintains a dynamic population at a maximum of $N$ instruction candidates, denoted as $\mathcal{P} = \{I_j\}_{j=1}^N$. Each instruction $I_j$ is associated with a learnable importance weight $w_j$, forming a set of weights $\mathcal{W}=\{w_j\}_{j=1}^N$ that correspondingly assigns the importance to each candidate.
During the training steps, for each question $q$, an instruction $I_j$ is sampled from the population $\mathcal{P}$ based on a selection probability distribution, $ I \sim p(\cdot|\mathcal{W; \mathcal{P}})$, conditioned on the importance weights $\mathcal{W}$. The selection probability for each instruction is calculated using a softmax function with a temperature parameter $\tau_s$ that controls the exploration of new instructions against the exploitation of known effective ones: 
\begin{equation}
    p(I_j | \mathcal{W};\mathcal{P}) = \frac{\exp(w_j / \tau_s)}{\sum_{j=1}^N \exp(w_j / \tau_s)}.
    \label{eq.softmax}
\end{equation}
It allocates higher probabilities for more effective instructions while down-weighting the proportion of less-performing instructions in sampling. This instruction sampling mechanism in \ours~enables us to integrate instruction optimization directly into the RL objective. The policy is similarly optimized by maximizing the expected reward, where the expectation is now taken over not only the distribution of questions $q\sim\mathcal{D}$ but also the distribution of instructions $ I \sim p(\cdot|\mathcal{W; \mathcal{P}})$. The learning objective of \ours~is thus extended as:
\begin{equation}
\begin{aligned}
\mathcal{J}_{\text{\ours}}(\theta, \mathcal{W}) = &\mathbb{E}_{q \sim \mathcal{D}, \textcolor{paleblue}{I \sim p(\cdot|\mathcal{W};\mathcal{P})},\, \{\tau_i\}_{i=1}^G \sim \pi_{\theta_{\text{old}}}(\cdot \mid \textcolor{paleblue}{I}, q; \mathcal{T})}\\
&\left[ \mathcal{L}_{\text{batch}}(\theta; \{\tau_i\}_{i=1}^G) \right]
\end{aligned},
\end{equation}
where $\mathcal{L}_{\text{batch}}$ denotes the bracketed term in Eq. (\ref{eq1}). The reward signals obtained from the samples now act for two objectives simultaneously: 1) updating the policy parameters $\theta$ via the \ours~policy gradient objective; 2) updating the importance weights $\mathcal{W}$ of the population of instructions. After a trajectory $\tau\sim \pi_{\theta_{t-1}}(\cdot | I_j, q; \mathcal{T})$ generated with instruction $I_j$ receives a reward $r(\tau)$, the reward can be automatically attributed back to $I_j$ based on the assignment. We then update its corresponding importance weight $w_j$ using a simple moving average of the mean reward $\bar{r}$ at a fixed window size of the most recent $n$ steps for a stable estimation of the importance over the group of trajectories generated by $I_j$:
\begin{equation} w_j^{(t)} = \frac{1}{n} \sum_{k=0}^{n-1} \bar{r}_{t-k, j}, \quad \text{where } \bar{r}_{t, j} = \frac{1}{G} \sum_{i=1}^{G} r(\tau_{i}). \end{equation}

To ensure a self-improving population of instructions, the population $\mathcal{P}$ is periodically pruned and evolved. At regular intervals, we apply a pruning phase for population control akin to the Successive Halving algorithm \citep{jamieson2016non}, where the bottom half of instruction candidates, ranked by their importance weights $w_j$, are pruned from the population, thereby focusing the policy optimization on more promising instructions. The top-performing candidates are then treated as parents to evolve new and potentially superior instructions. 

\begin{algorithm}[t]
\begin{footnotesize}
\caption{\ours: Instruction-Policy Co-Evolution}
	\label{alg:main_alg}
	\begin{algorithmic}[1]
		\STATE \textbf{Input}: Initial policy $\pi_{\theta_0}$, LLM-based optimizer $\mathcal{O}$, training steps $T$, population size $N_\mathcal{P}$, number of instructions as parents $N_\text{parent}$, pruning and evolving frequency $K_p$, $K_e$, evolving steps $T_e$.
		\STATE \textbf{Initialize}: Initial population $\mathcal{P}_0$, importance weights $\mathcal{W}_0 \leftarrow \{0\}^{N_{\mathcal{P}}}$, replay buffer $\mathcal{B} \leftarrow \emptyset$.
		\FOR{$t=1, \dots, T$}
        \STATE \textcolor{paleblue}{[Policy Optimization with Dynamic Instructions]}
		    \STATE Sample a question and instruction pair $q \sim \mathcal{D}$,  $I_j \sim p(\cdot | \mathcal{W}_{t-1}; \mathcal{P}_{t-1})$.
		    \STATE Generate trajectories $\{\tau_i\}_{i=1}^G \sim \pi_{\theta_{t-1}}(\cdot | I_j, q; \mathcal{T})$.
		    \STATE Compute rewards $\{r_i\}_{i=1}^G$ for each trajectory $\tau_i$.
		    \STATE Update $\theta_t \leftarrow \text{\ours.Update}(\theta_{t-1}, \{\tau_i\}, \{r_i\})$.
		    \STATE Update importance weight $w_j$ using $\bar{r} = \frac{1}{G}\sum r_i$.
            		    \IF{$t~(\mathrm{mod~}K_p) = 0$ \textbf{and} $|P_{t-1}|>N_{\text{parent}}$}
		        \STATE Rank instructions in $\mathcal{P}_{t-1}$ by weights $\mathcal{W}_{t-1}$.
		        \STATE Prune bottom $N/2$ instructions to create a new pool $\mathcal{P}_{t}$, s.t.  $|\mathcal{P}_{t}| \geq N_{\text{parent}}$.
		    \ENDIF
            \STATE \textcolor{palered}{[Experience-Driven Instruction Generation]}
            \STATE Add $(I_j, q, \tau_i, r_i)$ to the replay buffer $\mathcal{B}$.

		    \IF{$t~(\mathrm{mod~}K_e) = 0$ \textbf{and} $t < T_e$}
		        \STATE $\mathcal{P}_{\text{parent}} \leftarrow\text{top}_{N_{\text{parent}}}(\mathcal{P}_{t-1})$; $\mathcal{P}_{\text{t}}\leftarrow \emptyset$;  $\mathcal{W}_{\text{t}}\leftarrow \{0\}^{N_{\mathcal{P}}}$.
		        \WHILE{$|\mathcal{P}_{\text{t}}| < N_{\mathcal{P}}$}
		            \STATE Sample $I_{\text{parent}}$ from $\mathcal{P}_{\text{parent}}$ and failure trajectories $\{\tau_{\text{fail}}\}$ from $\mathcal{B}$.
		            \STATE Generate \{$I_{\text{new}}\} \sim \mathcal{O}(\cdot | I_{\text{parent}}, \{\tau_{\text{fail}}\})$.
		            \STATE Verify \{$I_{\text{new}}\}$ on a validation set $\mathcal{D}_{\mathcal{B}}$ and add the top-n candidates to $\mathcal{P}_{\text{new}}$.
		        \ENDWHILE
		        \STATE Set new population $\mathcal{P}_t \leftarrow \mathcal{P}_{\text{parent}} \cup \mathcal{P}_{\text{new}}$, s.t. $|\mathcal{P}_t|=N_\mathcal{P}$.
		    \ENDIF
		\ENDFOR
		\STATE \textbf{Return} optimized policy $\pi_{\theta_T}$ and best instruction $I^* = \text{argmax}_{I_j \in \mathcal{P}_T} w_j$.
	\end{algorithmic}
\end{footnotesize}
\end{algorithm}
\subsection{Experience-Driven Instruction Generation}
To generate novel and effective instructions, \ours~employs an on-policy reflection mechanism to generate new instructions. Inspired by the experience replay in RL \citep{lin1992self, mnih2015human}, we introduce a replay buffer $\mathcal{B}$ in \ours~for proposing new instructions via reflecting on past trajectories in memory, which contain interaction history with the environment. At each training step, we store a tuple of $(I, q, \tau, r(\tau))$ into the buffer $\mathcal{B}$. We further prioritize trajectories that resulted in incorrect or low-reward predictions from the latest step, creating a source of correction signals based on past failures. Periodically, once we reach a stable estimation of the importance weight, we start evolving new instruction candidates by the following on-policy reflection mechanism that involves three key steps:
\begin{enumerate}[leftmargin=5mm]
    \item \textbf{Select}: When the early phase has pruned the population to a target size, we select the top $N_\text{parent}$ instructions as the best-performing candidates to serve as parents for evolution. We randomly sample one instruction $I_t$ at a time to form the parent for the generation process.
    \item \textbf{Generate}: Together with the parent instruction, we sample a batch of failure trajectories from the replay buffer $\mathcal{B}$ that were generated in the latest step. These trajectories contain rich information that reveals the failure modes of the newest policy and its history of interaction with the environment. We then leverage an LLM-based optimizer $\mathcal{O}$ to reflect on failure patterns and propose new instruction candidates. Precisely, the optimizer is prompted with the parent instruction and the failure trajectories, and is tasked with performing self-reflection: it analyzes the causes of failures, critiques the existing strategy in instruction, and proposes a group of refined instructions $\{I_{\text{new}}\}$ designed to guide the agent to overcome these erroneous actions. This stands as a critical step for \ours~that constantly involves online feedback from the environment into the policy learning process to the agent. 
    \item \textbf{Verify:} Newly generated instructions are finally verified before merging into the active population to maintain the stability of the training.  Each proposed candidate $I_{\text{new}}$ is evaluated via a low-cost proxy, where we sample a subset of samples $\mathcal{D}_\mathcal{B}$ from the replay buffer for validation, and only the top-n best performing candidates are added to the new population $\mathcal{P}_t$. 
\end{enumerate}

This generate-and-verify loop continues until the population is refilled to the target population size $N_\mathcal{P}$, ensuring a self-improving population of instructions. In the inference time, we only return the best instruction with the highest importance weight $I^* = \text{argmax}_{I_j \in \mathcal{P}_T} w_j$, which is portable and transferable, with no additional overhead for LLM inference. The complete procedure for \ours~is provided in Algorithm~\ref{alg:main_alg}. Overall, the dynamic instruction sampling integrates natively with the policy optimization loop for exploring diverse reasoning strategies, and the experience-driven self-reflection mechanism enables the agent to constantly incorporate new environmental feedback into the policy learning process.

\begin{table*}[!t]
    \centering
        \caption{Main results on multi-hop and general question-answering benchmarks. All results are reported using exact match (EM) scores. We report baseline results from \citet{jin2025search}. We report the test performance with training steps in multiple random seeds in Fig.~\ref{fig:runs}}
        \resizebox{\textwidth}{!}{
        \renewcommand{\arraystretch}{1}
    \begin{tabular}{lcccccccc}
    \toprule
    \rowcolor{gray!20}
            \multicolumn{9}{c}{\texttt{Qwen-2.5-3B}} \\
\midrule
        Method & HotpotQA & 2WikiMQA & MuSiQue & Bamboogle & NQ & TriviaQA & PopQA & Avg. \\
        \cmidrule(lr){1-1} \cmidrule(lr){2-8} \cmidrule(lr){9-9}

        \rowcolor{gray!10}
\multicolumn{9}{c}{\textit{Model Inference}} \\
\midrule
        Direct & 0.149 & 0.244 & 0.020 & 0.024 & 0.106 &0.288 & 0.108 & 0.134 \\ 
        SFT & 0.186 & 0.248 & 0.044 & 0.112 & 0.249 & 0.292 & 0.104 & 0.176\\
        GRPO & 0.201 & 0.268 & 0.055 & 0.224 & 0.226 & 0.455 & 0.173 & 0.229 \\

        \midrule
                \rowcolor{gray!10}
\multicolumn{9}{c}{\textit{Tool-integrated Methods}} \\
\midrule
        IRCoT & 0.164 & 0.171 & 0.067 & 0.240 & 0.111 & 0.312 & 0.200 & 0.181 \\ 
        RAG & 0.255 & 0.226 & 0.047 & 0.080 & 0.348 & 0.544 & 0.387 & 0.270 \\
        SFT & 0.240 & 0.233 & 0.059 & 0.210 & 0.294 & 0.488 & 0.332 & 0.265 \\ 
        Search-o1 & 0.221 & 0.218 & 0.054 & 0.320 & 0.238 & 0.472 & 0.262  & 0.255\\
        \rowcolor{cyan!5}Search-R1 & 0.309 & 0.281 & 0.074 & 0.126 & \textbf{0.438} & \textbf{0.599} & 0.425 & 0.322\\ 
        \rowcolor{cyan!5}\textbf{\ours}~(ours) & \textbf{0.377} & \textbf{0.359} & \textbf{0.151} & \textbf{0.331} & 0.437 & 0.587 & \textbf{0.431} & \textbf{0.382}\\
        \midrule
        \rowcolor{gray!20}
        \multicolumn{9}{c}{\texttt{Qwen-2.5-7B}} \\
\midrule
        Method & HotpotQA & 2WikiMQA & MuSiQue & Bamboogle & NQ & TriviaQA & PopQA & Avg. \\
        \cmidrule(lr){1-1} \cmidrule(lr){2-8} \cmidrule(lr){9-9}

        \rowcolor{gray!10}
\multicolumn{9}{c}{\textit{Model Inference}} \\
\midrule
        Direct & 0.183 & 0.250 & 0.031 & 0.120 & 0.134 &0.408 & 0.140 & 0.181 \\ 
        SFT & 0.217 & 0.259 & 0.066 & 0.112 & 0.318 & 0.354 & 0.121 & 0.207\\
        GRPO & 0.242 & 0.273 & 0.083 & 0.296 & 0.297 & 0.539 & 0.202 & 0.276 \\

        \midrule
                \rowcolor{gray!10}
\multicolumn{9}{c}{\textit{Tool-integrated Methods}} \\
\midrule
        IRCoT & 0.133 & 0.149 & 0.072 & 0.224 & 0.224 & 0.478 & 0.301 & 0.239 \\ 
        RAG & 0.299 & 0.235 & 0.058 & 0.208 & 0.349 & 0.585 & 0.392 & 0.304 \\
        SFT & 0.331 & 0.296 & 0.123 & \textbf{0.355} & 0.360 & 0.592 & 0.380 & 0.348 \\ 
        Search-o1 & 0.187 & 0.176 & 0.058 & 0.296 & 0.151 & 0.443 & 0.131  & 0.206\\
        \rowcolor{cyan!5}Search-R1 & 0.356 & 0.308 & \textbf{0.128} & \textbf{0.355} & 0.423 & 0.575 & 0.411 & 0.365\\ 
        \rowcolor{cyan!5}\textbf{\ours}~(ours) & \textbf{0.368} & \textbf{0.349} & 0.115 & 0.339 & \textbf{0.476} & \textbf{0.646} & \textbf{0.455} & \textbf{0.393}\\
        \bottomrule
    \end{tabular}}
    \label{tab:main}
\end{table*}

\begin{figure*}[!t]
    \centering
    \includegraphics[width=1\linewidth]{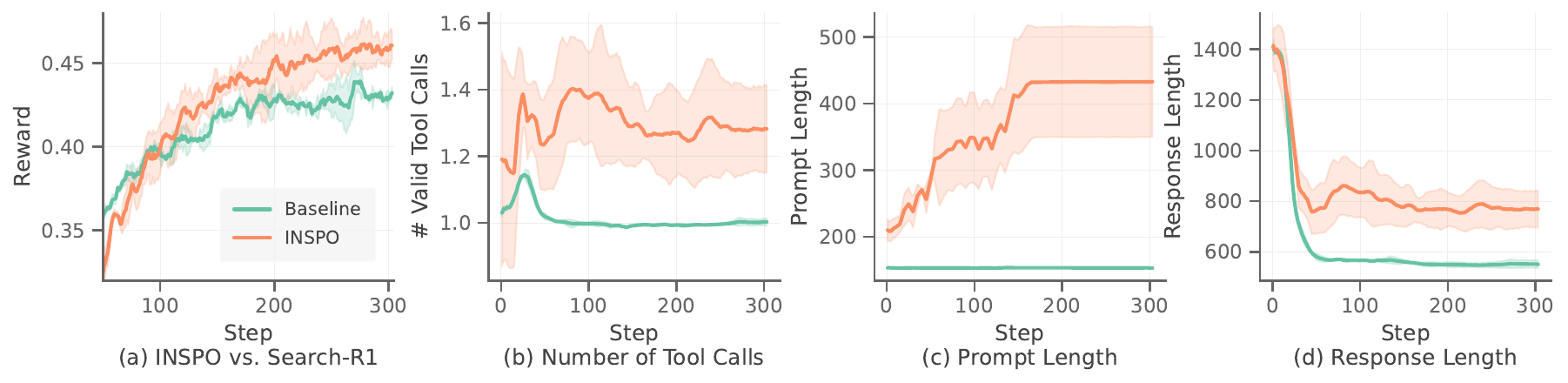}
    \vspace{-2mm}
    \caption{The training metric comparison between \ours~and the baseline (Search-R1) over three multiple random seeds, where the shaded area represents the 95\% confidence interval. (a) \ours~vs. Search-R1: \ours~shows a better reward at convergence compared to the Search-R1 baseline. (b) Number of tool calls: \ours~enables the agent to explore in a large regime of tool-use patterns (as shown by the large variance), and discovers instructions that lead agents to leverage a larger number of tool usages for solving the problem, whereas the baseline converges to a single-turn tool-use. (c) Prompt length: Periodically, \ours~evolves longer and more effective instructions along the RL training process, whereas the baseline sticks to a static instruction. (d) Response length: With a larger number of tool calls by \ours, the converged response comes with more tokens, which contains richer information from the search engine.}
    \label{fig:reward}
\end{figure*}

\section{Experiments}
\label{sec.exp}
\sparagraph{Models and evaluation data}
We evaluate our approach on a diverse set of tool-integrated question-answering benchmarks with the search tool. For multi-hop question answering, which requires complex reasoning and multi-turn information retrieval, we include HotpotQA \citep{yang-etal-2018-hotpotqa}, 2WikiMQA \citep{ho-etal-2020-constructing}, MuSiQue \citep{trivedi-etal-2022-musique}, and Bamboogle \citep{press-etal-2023-measuring}. For general question answering, we include Natural Questions (NQ) \citep{kwiatkowski-etal-2019-natural}, TriviaQA \citep{joshi-etal-2017-triviaqa}, and PopQA \citep{mallen-etal-2023-trust}.
We conduct \ours~experiments on Qwen 2.5 series \citep{qwen2025qwen25technicalreport}, specifically the 3B and 7B versions. We leverage the state-of-the-art reasoning model, Gemini 2.5 Pro \citep{comanici2025gemini}, as the LLM-based optimizer.

\begin{figure*}[!t]
    \centering
    \includegraphics[width=1\linewidth]{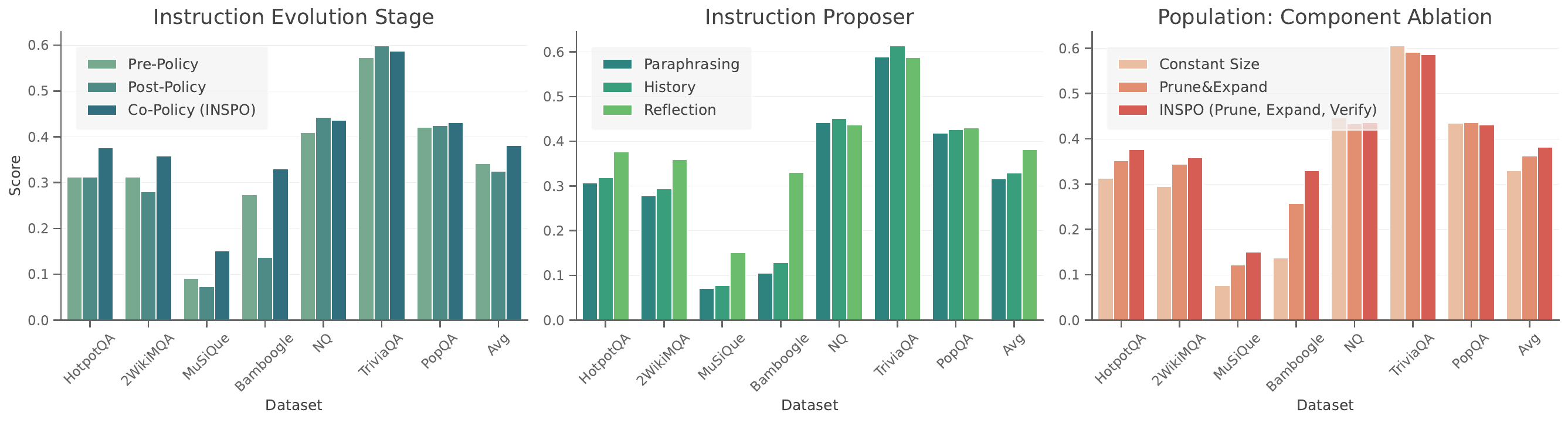}
    \vspace{-5mm}
    \caption{Ablation studies on the components of \ours~on Qwen-2.5-3B. The results show that (1) online co-evolution is superior to offline instruction optimization, (2) reflection-based proposers are more effective than simpler alternatives (prompts given in App. \S\ref{app.template}), and (3) all components of the co-evolution loop (Prune, Expand, and Verify) contribute to the final \ours~performance.}
    \label{fig:ablation}
\end{figure*}

\rparagraph{Baselines}
We compare \ours~against the following baselines that all come with a static instruction:
\begin{itemize}[leftmargin=5mm]
\item \textbf{Model Inference}: We include baselines without tool-using. We compare against the direct inference approach (Direct), Supervised Fine-Tuning (SFT) \citep{chung2024scaling}, and GRPO \citep{shao2024deepseekmath} on the same question-answering training data mixture without tool integration.
\item \textbf{Tool-Integrated Methods}: For tool-integrated reasoning, we include Retrieval-Augmented Generation (RAG) \citep{patrick2020rag}, IRCoT \citep{trivedi-etal-2023-interleaving}, Search-o1 \citep{li2025search}, SFT (reject sampling) \citep{jin2025search}, and the state-of-the-art Search-R1 \citep{jin2025search} approach that extends GRPO to multi-turn interleaved reasoning with a search engine.
\end{itemize}
\rparagraph{Setups}
For all tool-integrated experiments, we follow the experimental setup as Search-R1, and use the E5 \citep{wang2022text} as the retriever with a knowledge source on the 2018 Wikipedia dump \citep{karpukhin-etal-2020-dense}. The training data consists of a mixture of the NQ and HotpotQA training sets. We use the Exact Match (EM) as the reward function. For all inference-only baselines, we report the performance from \texttt{instruct} models, and the rest using the \texttt{base} models for training. In the \ours~framework, we maintain a population size $N_\mathcal{P}$ of 7, $N_\text{parent}$ of 1 for exploitation, and a temperature $\tau_s$ of 0.2 in computing the selection probability for balancing the exploration and exploitation on candidates. We set a pruning period $K_p$ at 5 and the evolution period $K_e$ at 15. We evolve $T_e=150$ steps out of a total of 300 steps, thereby allowing the policy to be sufficiently optimized with the \ours-evolved instruction. We refer readers to App.~\S \ref{app.details} and Fig.~\ref{fig:hyperparameter} for details.
\section{Results}
\sparagraph{Main results}
We present the main results of \ours~compared to baselines in Table~\ref{tab:main}. \ours~yields substantial performance improvements over model inference and tool-integrated baselines that all come with a static instruction. \ours~achieves an average EM score of 38.2\% on all benchmarks with Qwen-2.5-3B, surpassing the state-of-the-art RL baseline with the search tool, Search-R1, by a large margin of 6\%. This performance advantage is consistent when scaling to the 7B model. In particular, we observe that \ours~excelled in complex multi-hop reasoning tasks, such as HotpotQA and 2WikiMQA, where \ours~outperforms the Search-R1 baseline over 7\% in the 3B model. This is further reflected in Fig.~\ref{fig:reward} (b), where the co-evolution of instruction and policy enables the agent to explore and execute in a regime of multiple valid tool calls, thereby learning to utilize more turns of interactions with the environment to resolve more challenging questions. In contrast, the baseline, constrained by a static instruction, converges to a policy with a mean of tool calls that is marginally above $1$, limiting its ability to decompose and solve multi-hop questions. In addition, we observe that \ours~periodically evolves to longer prompt length as shown in Fig.~\ref{fig:reward} (c). It validates the functionality of the self-reflection mechanism in instruction generation, where \ours~constantly involves environmental feedback to refine the instruction with new details. These results collectively show that by treating the instruction as a dynamic and optimizable component of the learning loop, \ours~unlocks more effective and strategic tool-integrated reasoning paths.

\begin{figure*}[!t]
    \centering
    \includegraphics[width=1\linewidth]{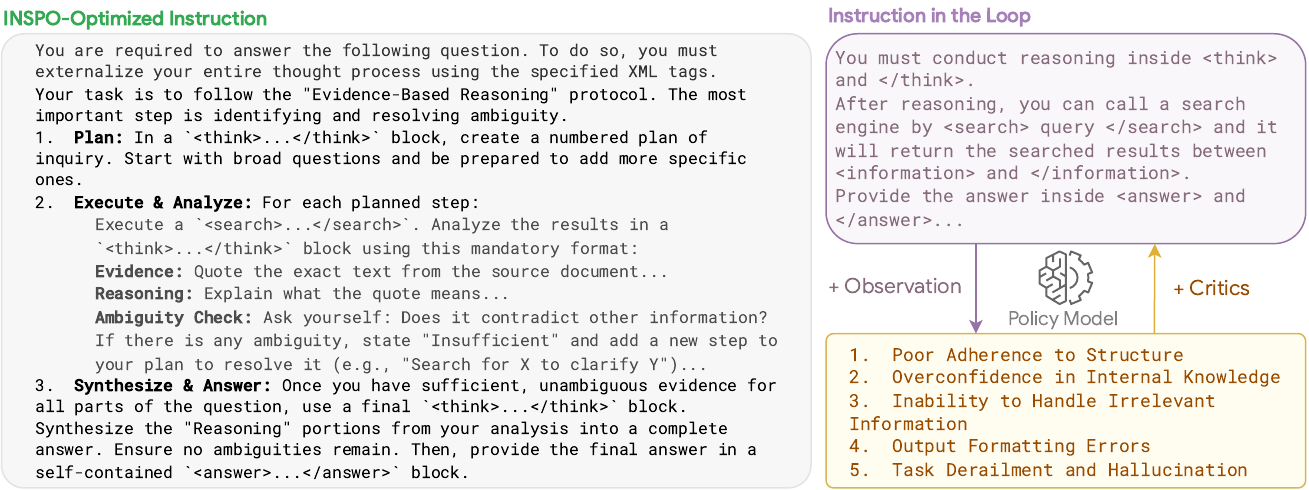}
    \vspace{-2mm}
    \caption{A demonstration of the instruction co-evolution process with the policy model for using the search tool. The policy model is first prompted with the instruction-question pair, generating trajectories with environmental feedback and rewards, which are passed to an LLM-based optimizer for the experience-driven reflection process. The optimizer generates critiques on the failures and proposes new instruction candidates, forming an online optimization loop for the instruction.}
    \label{fig:demo}
\end{figure*}

\rparagraph{Case analysis}
We further inspect the agent trajectory from both baseline and \ours~to better understand the source of the performance disparity. In App.~\S\ref{app.demo}, we observe that the baseline model, trained with a static instruction, attempts to solve the question with a single and broad search query that mirrors the input question. It fails to retrieve the necessary evidence, leading the agent to hallucinate an incorrect final answer. In Fig.~\ref{fig:demo} and App.~\S\ref{app.demo}, we show an example of the instruction evolution loop that provides insight into how \ours~evolves better instructions. By analyzing trajectories given the initial instruction, the LLM-based optimizer in \ours~identifies common failure modes, such as ``\textit{poor adherence to structure}" and  ``\textit{overconfidence in internal knowledge}". It then generates new instruction candidates that explicitly mitigate these specific flaws by, for instance, mandating a step-by-step plan for tool-using and enforcing a strict \texttt{XML} format for actions. The new instruction reflects on prior failures, enriches the characteristics, and forces the agent to follow a step-by-step plan. This structured reasoning directly leads to the correct answer.

\rparagraph{Instruction Evolution Stage}
We compare our online co-evolution with two offline APO alternatives: pre-policy optimization, where the instruction is optimized \textit{before} RL training begins, and post-policy optimization, where the instruction is tuned \textit{after} the training is complete. We observe that only the pre-policy brings subtle gains to the converged performance, whereas the post-policy model has become overfitted to the initial instruction, and any new instruction candidates only degrade the performance. It validates that the optimal instruction changes as the policy learns, and co-evolution is crucial for achieving the best performance.

\rparagraph{Instruction Proposer}
We replace our reflection-based optimizer with two simpler APO alternatives to ablate the effectiveness of the reflection mechanism. In Fig.~\ref{fig:ablation}, we include a paraphrasing-based optimizer \citep{zhou2023large} that generates $I\sim\mathcal{O}(\cdot|I_\text{old})$ and a history-based \citep{yang2024large} prompt optimizer with $I\sim\mathcal{O}(\cdot|\{I_i,r_i\}_{i=1}^n )$ that conditions on instruction-reward pairs. It shows that both optimizers are inadequate to yield meaningful gains along the policy optimization. It underscores the importance of the self-reflection mechanism that incorporates online feedback to improve the agent learning process.

\rparagraph{Instruction Population}
Finally, we ablate the core modules in \ours. We start from the \ours~with a constant population size without any pruning, which shows an average score of 33.0\%. Applying the periodic pruning of low-performing candidates and expanding with high-performing ones improves \ours~performance to 36.3\%, and the full \ours~approach that includes the complete generate-and-verify loop achieves the best performance. This highlights the importance of maintaining a high-quality population of instructions, which can drive the policy models towards better policy and generate richer reward signals.

\section{Related Work}
\sparagraph{Reinforcement Learning for LLMs}
Reinforcement learning (RL) \citep{sutton1999reinforcement} has been a powerful paradigm for post-training LLMs, notably in aligning LLMs with human preference \citep{ouyang2022training, rafailov2023direct}. It typically trains a reward model and optimizes the policy via Proximal Policy Optimization (PPO) \citep{schulman2017proximal}. Recently, DeepSeek-R1 \citep{guo2025deepseek} has demonstrated the effectiveness of using rule-based verifiable rewards (RLVR) in advancing the reasoning capabilities of LLMs, where GRPO \citep{shao2024deepseekmath} obviates the need for a critic model by group-wise sampling. DAPO \citep{yu2025dapo} further introduces techniques that improve the training stabilities of RLVR, such as clipping-higher and dynamic sampling. In addition, Dr.~GRPO \citep{liu2025understanding} rectifies the length bias in GRPO. Orthogonal to these approaches, \ours~serves as a modular component to the typical RLVR learning loop.

\rparagraph{Tool-Integrated Agents}
LLMs can be instructed to follow the pattern of reasoning and acting, which enables agentic applications with tool integration \citep{yao2023react}. For instance, IRCoT \citep{trivedi-etal-2023-interleaving} interleaved chain-of-thought reasoning \citep{wei2022chain} with information retrieval. Toolformer \citep{schick2023toolformer} trains LLMs with diverse tool usages via SFT. More recently, the RL paradigm has led to agents with multi-turn tool interactions \citep{feng2025retool, li2025torl}, where the search tool has excelled in resolving question-answering tasks \citep{jin2025search, song2025r1}. Though there is a growth of end-to-end RL frameworks for multi-turn tool-use \citep{xue2025simpletir, jiang2025verltool}, the importance of instruction design for agentic policy optimization has been largely overlooked, and they all rely on a costly manual design for the instruction that initializes the interaction with tools while treating it as a static component along the learning process.

\rparagraph{Instruction Optimization}
The instruction serves as the protocol for defining the agents' behaviors, where language agents have been exposed to sensitivity with respect to prompt designs \citep{zhou2025multi}. Traditional instruction optimization (IO) approaches rely on simply paraphrasing the instruction \citep{zhou2023large, zhou-etal-2024-fairer}. \citet{2024ioes, wan2025from} emphasizes the importance of demonstration/exemplars together with the general instruction design. Recent approaches integrate LLM-based optimizers \citep{yang2024large} and leverage textual gradients \citep{pryzant-etal-2023-automatic, 2025textgrad} to optimize the instruction, where \citet{2025agrawal} show that reflective prompt design can even outperform the RL paradigm. Previously, \citet{soylu-etal-2024-fine} propose interleaving IO with SFT. In the pioneering context of RL, different from traditional IO approaches that optimize the instruction before commencing the RL training \citep{zhang2025merf}, \ours~is the first framework to integrate IO directly into the online RL loop, co-evolving instructions and the policy model with on-policy experience and environmental feedback.

\section{Conclusion}
In this work, we introduced \ours, a novel framework that enables the co-evolution of instructions and policy learning for agentic policy optimization. Different to existing RLVR approaches that rely on static, manually-designed instructions, \ours~maintains a dynamic population of instruction candidates and employs an experience-driven instruction generation mechanism, automating the discovery of more effective reasoning strategies via reflecting on online feedback along the policy learning process. Our experiments demonstrate that \ours~substantially outperforms strong baselines with static instructions on a wide range of tool-integrated question-answering benchmarks. The case studies and ablation analyses further confirmed that the online, reflection-based co-evolution of instructions is key to guiding the agent toward more robust and accurate tool-integrated reasoning. \ours~represents a significant step toward creating a more autonomous and adaptive RL framework, reducing the burden of manual prompt engineering and unlocking new capabilities in LLM agents.

\section*{Acknowledgements}
The work has been supported by the UK Research and Innovation (UKRI) Frontier Research Grant EP/Y031350/1 (the UK government’s funding guarantee for ERC Advanced Grants) awarded to Anna Korhonen at the University of Cambridge. The work has also been supported in part by a Royal Society University Research Fellowship (no 221137; 2022-) awarded to Ivan Vuli\'{c}, and by the UK EPSRC grant EP/T02450X/1.
\bibliography{icml2026}
\bibliographystyle{icml2026}

\newpage
\appendix
\onecolumn
\section{Future Work and Limitations}
While \ours~demonstrates substantial performance improvement, the effectiveness of our experience-driven instruction generation relies on a capable LLM-based optimizer for self-reflection and generating sophisticated instructions. We have primarily studied using Gemini 2.5 Pro as the optimizer, and we will endeavor to explore other thinking-style models as part of our future work. We would like to note that \ours~still introduces a marginal but non-zero computation overhead that includes the API costs for the LLM-based optimizer and the inference cost for the phase of verification, which we show is marginal compared to the cost of the total training in Appendix.~\ref{app.details}. Lastly, our current experiments have primarily focused on search-based question answering tasks. We believe the core principles of \ours~are broadly applicable to other agentic domains, and we will explore extending the framework with other available tools in future work. 

\section{Implementation Details}
\label{app.details}
We build our codebase on the VeRL framework \citep{sheng2025hybridflow}. Our experimental setup, including the training and evaluation data mixture, follows that of Search-R1 \citep{jin2025search}. 
We conduct our training on a node of 8$\times$H100 GPUs. For all retrieval-based methods, we keep the top-3 retrieved passages from the retriever. We used a learning rate of 5e-7 for the \texttt{Qwen2.5-3B-base} model and 3e-7 for \texttt{Qwen2.5-7B-base} model. All methods were trained for a total of 300 steps with a cosine learning rate schedule with a warm-up ratio of 0.1, which were then evaluated every 100 steps, and we report the final checkpoint before any reward collapse.

For both GRPO and \ours~training. We set the group size $G$ to 5. We used a batch size of 512, a mini-batch size of 256, and a micro-batch size of 64. We set the maximum prompt length to 4096 and truncate response and observation lengths to 500 for each turn. The maximum number of turns was capped at 4 for the \texttt{Qwen2.5-3B-base model} and 6 for \texttt{Qwen2.5-7B-base} as stronger models tend to leverage more turns of tool-use. For LLM inference, we used a temperature of 1.0 and top-p of 1.0. We set the KL divergence coefficient $\beta$ of 0.001 and the clip ratio $\epsilon$ of 0.2.

In \ours-specific setting, our experiments initialize an initial set of diverse population with the original instruction from Search-R1 (see App.~\S\ref{app.demo}). The importance weight for each instruction candidate was updated using a running average over a window size of 5 steps. In the phase of experience-driven instruction generation, we randomly sampled 4 failure trajectories $\{\tau_{\text{fail}}\}$ from the replay buffer for self-reflection. For verification, we set $|\mathcal{D}_{\mathcal{B}}|$ of 200 samples as a low-cost proxy for validating the quality of instruction candidates, where we iteratively merge the best-performing candidates into the active population with the parent instructions. 
\section{Computation Analysis}
In terms of the computation overhead on top of the original RL training process, given $150/15=10$ times of evolution plus the initialization of the population, the self-reflection mechanism incurs roughly 11 API calls to the LLM-based optimizer. The phase of verification incurs roughly $6\times200
\times10\text{ }(N\times|\mathcal{D}_{\mathcal{B}}|\times T_e/K_e)=12000$ additional inference from the policy model, given 6 new candidates per evolution and a maximum of 5 rounds of verification retrials, which is only about 1.4\%-7\% compared to the total inference calls, $169615\times5=848075$ ($|\mathcal{D}|\times G$) with a group size of 5, for one epoch training.

\section{Additional Experiments}
\begin{figure*}
    \centering
    \includegraphics[width=1\linewidth]{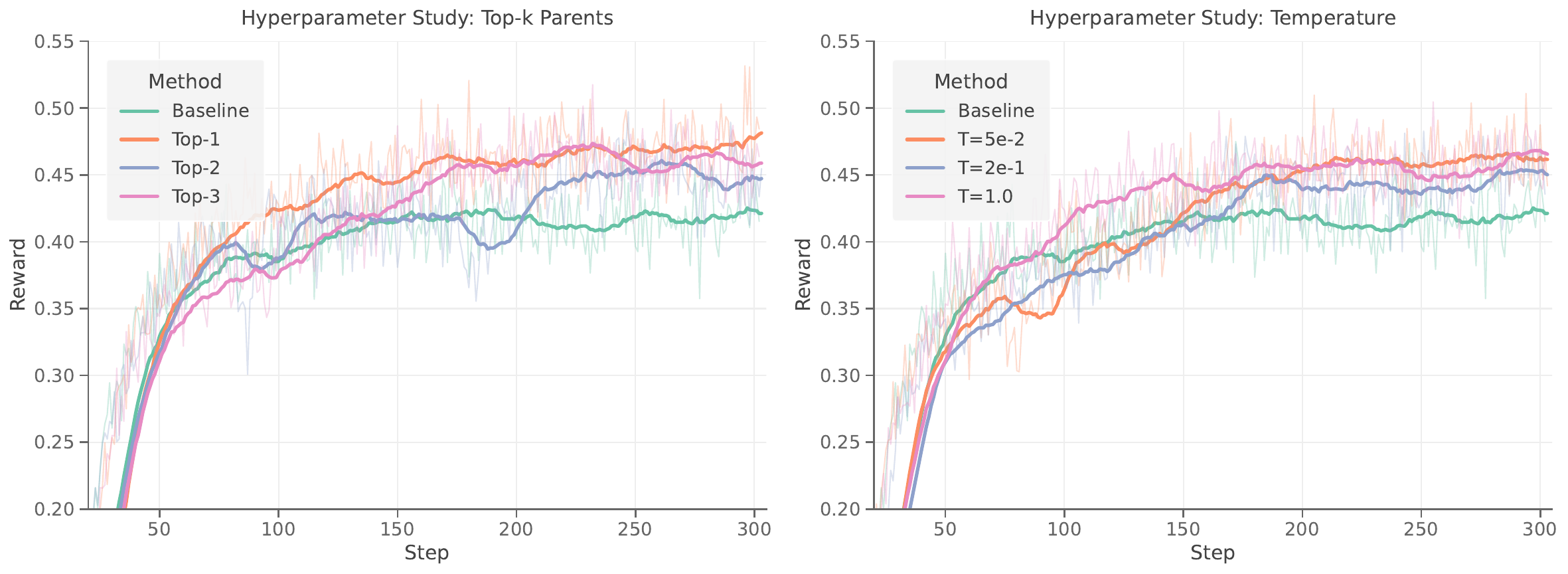}
    \vspace{-5mm}
    \caption{The hyperparameter study of \ours~with a reference baseline performance from the Search-R1 approach. (a) We study using the top-$N_{\text{parents}}$ for evolving new instruction candidates. (b) Ablating the temperature term $\tau_s$ in Eq.~(\ref{eq.softmax}).}
    \label{fig:hyperparameter}
\end{figure*}

In Fig.~\ref{fig:hyperparameter}, we conducted a hyperparameter study in terms of both the top-$N_{\text{parents}}$ in evolving new instruction candidates and the temperature $\tau_s$ used in controlling the balance between exploration and exploitation of \ours~in Eq.~(\ref{eq.softmax}). It is observable that all \ours~curves with top-$N_{\text{parents}}$ consistently outperform the Search-R1 baseline. In particular, restricting the instruction generation to be conditioned only on the top-1 parent ($N_{\text{parents}}=1$) yielded the highest performance, suggesting that prioritizing the single highest-quality instruction as the parent is more effective than retaining multiple instruction candidates as parents for evolvement, and it can lead to a more greedy improvement for the population of instructions. In the temperature study, we observe that all temperature settings (0.05, 0.2, and 1.0) mainly affect the variance in the rewards (see the smoothness of the curve), which aligns with our intuition that a lower temperature comes with a more greedy exploitation on the existing set of instructions. Consequently, the sudden update of instructions can result in high variance in the following steps. Hence, we follow the mid-temperature option ($\tau_s=0.2$) for achieving the best balance between exploitation and exploration of instruction candidates. 

\begin{figure*}
    \centering
    \includegraphics[width=1\linewidth]{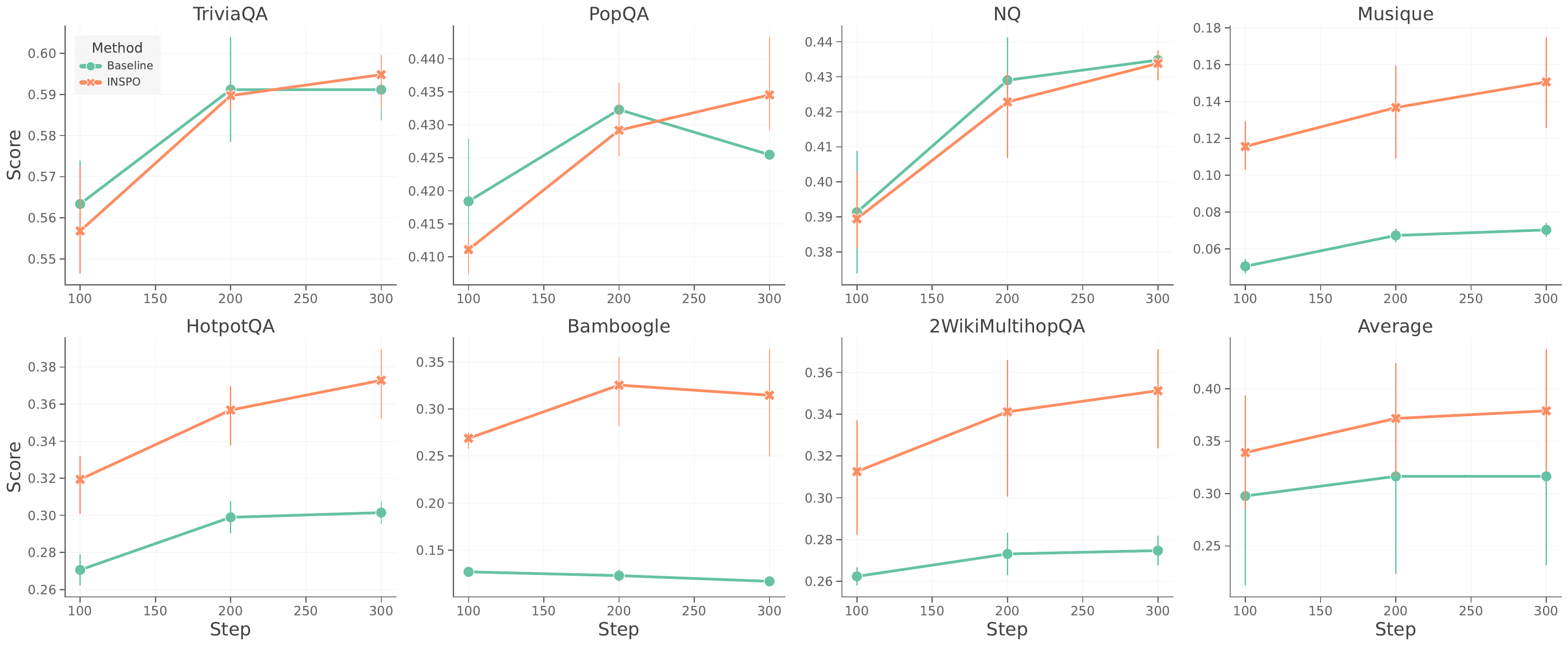}
    \vspace{-5mm}
    \caption{Test performance of \ours~against the Search-R1 baselines over 3 random seeds on Qwen-2.5-3B.}
    \label{fig:runs}
\end{figure*}
\clearpage
\section{Prompt Templates}
\label{app.template}
\begin{tcolorbox}[colback=blue!15!white, colframe=blue!50!white, coltitle=black, fonttitle=\bfseries, title=Paraphrasing-Based Instruction Optimizer, boxrule=0.5mm, arc=2mm, outer arc=2mm]
\footnotesize
\texttt{<ins\_0>}
\texttt{\{Instruction\}}
\texttt{</ins\_0>}

Improve the above prompt. 

Give 6 different candidates. 

Make sure you include the same XML tags \texttt{<>}...\texttt{</>} as the original prompt in your improved prompts.

For each XML tag, include both the beginning and ending tags, such as \texttt{<answer>}...\texttt{</answer>}.

List the instruction candidates in \texttt{<ins\_1>}...\texttt{</ins\_1>} to \texttt{<ins\_N>}...\texttt{</ins\_N>} tags.
\end{tcolorbox}

\begin{tcolorbox}[colback=green!15!white, colframe=green!70!black, coltitle=black, fonttitle=\bfseries, title=History-Based Instruction Optimizer, boxrule=0.5mm, arc=2mm, outer arc=2mm]
\footnotesize
\texttt{<history>}

\texttt{<ins\_i>}
\texttt{\{Instruction i\}}
\texttt{</ins\_i>}

\texttt{<score\_i>}
\texttt{\{Score i\}}
\texttt{</score\_i>}

...

\texttt{</history>}

Based on the above ``instruction: score" history for instructions with their scores, where the score ranges from 0 to 100, analyze what components can lead to better performance.

Give \texttt{N} different but better candidates. Make it concise and effective. Be creative and diverse in your suggestions.

Make sure you include the same XML tags \texttt{<>}...\texttt{</>} as the original prompt in your improved prompts.

For each XML tag, include both the beginning and ending tags, such as \texttt{<answer>}...\texttt{</answer>}.

List the instruction candidates in \texttt{<ins\_1>}...\texttt{</ins\_1>} to \texttt{<ins\_N>}...\texttt{</ins\_N>} tags.
\end{tcolorbox}

\begin{tcolorbox}[colback=mainboxbg, colframe=mainboxborder, coltitle=black, fonttitle=\bfseries, title=Reflection-Based Instruction Optimizer (\ours), boxrule=0.5mm, arc=2mm, outer arc=2mm]
\footnotesize

The initial prompt is:

\texttt{<ins\_0>}

\texttt{\{Instruction\}}

\texttt{</ins\_0>}

However, it still gets the following examples wrong:

\texttt{<example>}

Response: \texttt{\{Rollouts from Replay Buffer\}}

Correct answer: \texttt{\{Ground Truth\}}

...

\texttt{</example>}

Analyze the reasons behind these mistakes and why the prompt could have gotten these examples wrong.

Now take the reasons into account. Revise and improve the initial prompt by giving \texttt{N} different but better candidates. 

Make sure you include the same XML tags \texttt{<>}...\texttt{</>} as the original prompt in your improved prompts.

For each XML tag, include both the beginning and ending tags, such as \texttt{<answer>}...\texttt{</answer>}.

List the instruction candidates in \texttt{<ins\_1>}...\texttt{</ins\_1>} to \texttt{<ins\_N>}...\texttt{</ins\_N>} tags.
\end{tcolorbox}
\section{Demonstration}
\label{app.demo}
Response from the baseline. Incorrect Answer (Search-R1):
\begin{lstlisting}[style=mystyle]
Answer the given question. You must conduct reasoning inside <think> and </think> first every time you get new information. After reasoning, if you find you lack some knowledge, you can call a search engine by <search> query </search> and it will return the top searched results between <information> and </information>. You can search as many times as your want. If you find no further external knowledge needed, you can directly provide the answer inside <answer> and </answer>, without detailed illustrations. For example, <answer> Beijing </answer>.

Question: Who was president of the United States in the year that Citibank was founded?
<|im_end|>
<|im_start|>assistant
To pursue the concern correctly,

<search> Who was president of the United States in the year that Citibank was founded </search>

<information>Doc 1(Title: Citibank) resulting in a net profit for the U.S. federal government. The City Bank of New York was founded on June 16, 1812. The first president of the City Bank was the statesman and retired Colonel, Samuel Osgood. In August 1813, with Osgood's death, William Few became President of the Bank, staying until 1817, followed by Peter Stagg (1817-1825), Thomas Smith (1825-1827), Isaac Wright (1827-1832), and Thomas Bloodgood (1832-1843). Ownership and management of the bank was taken over by Moses Taylor in 1837...

Doc 2(Title: Citibank) century. During Taylor's ascendancy, the bank functioned largely as a treasury and finance center for Taylor's own extensive business empire. Later presidents of the bank included Gorham Worth (1843-1856), Moses Taylor himself (1856-1882), Taylor's son-in-law Patrick Pyne, and James Stillman (1891-1909). In 1831 City Bank was the site of one of America's first bank heists when two thieves made off with tens of thousands of dollars' worth of bank notes, and 398 gold doubloons. The bank also has the distinguishable history of financing war bonds for the war of 1812, serving as a founding member of the financial clearing house...

Doc 3(Title: Citigroup) Brothers in 1910. City Bank of New York was chartered by New York State on June 16, 1812, with $2 million of capital. Serving a group of New York merchants, the bank opened for business on September 14 of that year, and Samuel Osgood was elected as the first President of the company. The company's name was changed to The National City Bank of New York in 1865 after it joined the new U.S. national banking system, and it became the largest American bank by 1883 </information>

<answer> James Buchanan </answer>
\end{lstlisting}

\ours-optimized response. Correct Answer (\ours):
\begin{lstlisting}[style=mystyle]
Answer the given question. You must act as a meticulous researcher.
First, conduct your reasoning inside `<think>` and `</think>`. Create a step-by-step plan to identify all the key entities (people, places, events, objects) in the question.
You must use `<search>query</search>` to find information about each entity individually. Never search for the entire question. Search results will appear between `<information>` and `</information>`.
After each search, you must pause and analyze the results in a new `<think>` block. In your analysis, focus on the **context** of the information. Ask yourself:
*   Is this information current or historical?
*   Does this refer to a real-world fact or a detail from a fictional work?
*   Is this name/place/date definitive, or are there other possibilities?
State the key fact and its context clearly. Then, decide if another search is needed.
Once you have gathered sufficient verified and contextualized facts, you must perform a final synthesis in a concluding `<think>` block. This synthesis must:
1.  List the key facts and their important context (e.g., "Munich - filming location, not fictional setting").
2.  Logically connect these facts, respecting their context, to formulate an answer.
3.  Explicitly state the answer you have formulated based on this logic.
Finally, provide that direct answer inside `<answer>` and `</answer>`. Do not include illustrations or justifications in the answer. For example: `<answer>Beijing</answer>`.

Question: Who was president of the United States in the year that Citibank was founded?

To answer the question "Who was president of the United States in the year that Citibank was founded?", I'll follow these steps:

<think>
First, I'll identify the key entities in the question:
1. Citibank
2. Year of foundation
3. President of the United States at that time
</think>

<search>Who founded Citibank and what year was it established</search>

<information>Doc 1(Title: Citibank) resulting in a net profit for the U.S. federal government. The City Bank of New York was founded on June 16, 1812. The first president of the City Bank was the statesman and retired Colonel, Samuel Osgood. In August 1813, with Osgood's death, William Few became President of the Bank, staying until 1817, followed by Peter Stagg (1817-1825), Thomas Smith (1825-1827), Isaac Wright (1827-1832), and Thomas Bloodgood (1832-1843). Ownership and management of the bank was taken over by Moses Taylor in 1837...

Doc 2(Title: Citigroup) was formed on October 8, 1998, following the $140 billion merger of Citicorp and Travelers Group to create the world's largest financial services organization. The history of the company is, thus, divided into the workings of several firms that over time amalgamated into Citicorp, a multinational banking corporation operating in more than 100 countries; or Travelers Group, whose businesses covered credit services, consumer finance, brokerage, and insurance. As such, the company history dates back to the founding of: the City Bank of New York (later Citibank) in 1812; Bank Handlowy in 1870; Smith Barney in 1873, Banamex in 1884; Salomon...

Doc 3(Title: Citigroup) Brothers in 1910. City Bank of New York was chartered by New York State on June 16, 1812, with $2 million of capital. Serving a group of New York merchants, the bank opened for business on September 14 of that year, and Samuel Osgood was elected as the first President of the company. The company's name was changed to The National City Bank of New York in 1865 after it joined the new U.S. national banking system, and it became the largest American bank by 1880. By 1912, it started to branch...
</information>

<think>
From the search results, I can see that Citibank was founded in 1812.
</think>

<search>What was the president of the United States in 1812</search>

<information>Doc 1(Title: "1812 United States presidential election") 1812 United States presidential election The United States presidential election of 1812, the seventh quadrennial American presidential election, was held from Friday, October 30, 1812 to Wednesday, December 2, 1812. Taking place in the shadow of the War of 1812, incumbent Democratic-Republican President James Madison defeated DeWitt Clinton, who drew support from dissident Democratic-Republicans in the North as well as Federalists. It was the first presidential election to be held during a major war involving the United States. Northern Democratic-Republicans had long been dissatisfied by the Southern dominance of their party, and DeWitt Clinton's uncle, Vice President George Clinton, had...

Doc 2(Title: "War of 1812") sailors into the Royal Navy. Incidents such as the ""Chesapeake-Leopard"" affair, which happened five years before the war, inflamed anti-British sentiment in the US. In 1811, the British were in turn outraged by the ""Little Belt"" affair, in which 11 British sailors died. Britain supplied Native Americans who raided American settlers on the frontier, hindering American expansion and provoking resentment. Historians debate whether the desire to annex some or all of British North America (Canada) contributed to the American decision to go to war. On June 18, 1812, US President James Madison, after heavy pressure from the War Hawks in...

Doc 3(Title: "1812 United States elections") 1812 United States elections The 1812 United States elections elected the members of the 13th United States Congress. The election took place during the First Party System, and shortly after the start of the War of 1812. The Federalist Party made a relatively strong showing, winning seats in both chambers while supporting a competitive challenge to the incumbent Democratic-Republican President. However, the Democratic-Republican Party continued its control of the Presidency and both houses of Congress. In the Presidential election, incumbent Democratic-Republican President James Madison defeated New York Lieutenant Governor and New York City Mayor DeWitt Clinton. Clinton was a member</information>

<think>
From the search results, I can see that the president of the United States in 1812 was James Madison.
</think>

<answer>James Madison</answer>
\end{lstlisting}

A demonstration of reflection-based instruction optimizer (\ours):
\begin{lstlisting}[style=mystyle]
An analysis of the provided examples reveals several key reasons for the model\'s failures:

1.  **Poor Adherence to Structure:** The model frequently fails to use the `<think>` tag for its reasoning, instead writing conversational text or placing reasoning outside the designated tags. This suggests the instruction "you must conduct reasoning inside `<think>`" is not strong enough.
2.  **Overconfidence in Internal Knowledge:** In the "phone code" example, the model provided a wrong answer from its internal knowledge without attempting to verify it with a search. The prompt does not sufficiently discourage this behavior.
3.  **Inability to Handle Irrelevant Information:** In the "Dukes of Hazzard" example, the search results were not about the specific episode. The model failed to recognize this, could not formulate a new search query, and instead provided a nonsensical answer based on misinterpreting the result format ("Doc 1"). The prompt lacks instructions for self-correction.
4.  **Output Formatting Errors:** The model often includes conversational text like "Answer:" before the final answer or provides a full sentence instead of the concise answer requested by the example `<answer> Beijing </answer>`.
5.  **Task Derailment and Hallucination:** In the "Green Revolution" example, the model completely loses focus, starts discussing an unrelated topic, and provides a final answer (`A-Train`) that has no connection to the original question or the relevant search results.

The initial prompt is too permissive and lacks a clear, enforced workflow. To improve performance, the prompt needs to be more prescriptive, explicitly define a step-by-step process, mandate searching over internal knowledge, and include instructions for handling failed searches.

Here are six improved prompt candidates designed to address these issues.
<ins_1>
You must follow these steps in order to answer the question:
Step 1: First, and for every new piece of information, you must reason about the user\'s question inside <think> and </think> tags. In your reasoning, break down the question and create a plan.
Step 2: If your plan requires information you do not have, execute a search using the <search> query </search> tag.
Step 3: The search results will be provided in <information> and </information> tags. You must then reason about the results in a new <think> and </think> block to determine if you have the answer or need more information.
Step 4: Repeat steps 2 and 3 as many times as necessary.
Step 5: Once you are confident you have the final answer, provide it directly inside <answer> and </answer> tags. Do not add any other text or explanation outside the tags.
</ins_1>
...
\end{lstlisting}
\end{document}